\documentclass[10pt,twocolumn,letterpaper]{article}

\usepackage{wacv}
\usepackage{times}
\usepackage{epsfig}
\usepackage{graphicx}
\usepackage{amsmath}
\usepackage{amssymb}

\usepackage{booktabs}
\usepackage[ruled,linesnumbered]{algorithm2e}
\usepackage{lipsum}
\usepackage{kotex}
\usepackage{multirow}
\usepackage{mathtools}
\usepackage[noend]{algpseudocode}

\usepackage{enumitem}
\usepackage[table]{xcolor}
\usepackage{subcaption}
\usepackage{caption}
\usepackage[flushleft]{threeparttable} 
\usepackage{parcolumns}
\usepackage{pifont}
\definecolor{brickred}{rgb}{0.8, 0.25, 0.33}
\definecolor{brickred2}{rgb}{0.25, 0.8, 0.33}
\newcommand{\cm}{\color{brickred2}{\ding{51}}}%
\newcommand{\xm}{\color{brickred}{\ding{55}}}%
\def\Equal{\texttt{=}}
\newcommand*\samethanks[1][\value{footnote}]{\footnotemark[#1]}

\usepackage[accsupp]{axessibility}  

\def\wacvPaperID{902} 

\wacvfinalcopy 

\ifwacvfinal
\fi

\ifwacvfinal
\usepackage[breaklinks=true,bookmarks=false]{hyperref}
\else
\usepackage[pagebackref=true,breaklinks=true,colorlinks,bookmarks=false]{hyperref}
\fi

\begin{document}

\title{DAQ: Channel-Wise Distribution-Aware Quantization for Deep Image Super-Resolution Networks}
\author{Cheeun Hong\thanks{equal contribution}\\
\and
Heewon Kim\samethanks\\
\and
Sungyong Baik\\
\and
Junghun Oh\\
\and
Kyoung Mu Lee \vspace{0.1cm}\\
\hspace{-13cm}{Department of ECE, ASRI, Seoul National University}\\
\hspace{-13cm}{\tt\small \{cheeun914, ghimhw, dsybaik, dh6dh, kyoungmu\}@snu.ac.kr}\\
}

\maketitle

\ifwacvfinal
\thispagestyle{empty}
\fi

\begin{abstract}
   Since the resurgence of deep neural networks (DNNs), image super-resolution (SR) has recently seen a huge progress in improving the quality of low resolution images, however at the great cost of computations and resources.
   Recently, there has been several efforts to make DNNs more efficient via quantization.
   However, SR demands pixel-level accuracy in the system, it is more difficult to perform quantization without significantly sacrificing SR performance.
   To this end, we introduce a new ultra-low precision yet effective quantization approach specifically designed for SR.
   In particular, we observe that in recent SR networks, each channel has different distribution characteristics.
   Thus we propose a channel-wise distribution-aware quantization scheme. 
   Experimental results demonstrate that our proposed quantization, dubbed Distribution-Aware Quantization (DAQ), manages to greatly reduce the computational and resource costs without the significant sacrifice in SR performance, compared to other quantization methods.
    
\end{abstract}

\section{Introduction}
Image super-resolution (SR), one of the fundamental computer vision tasks, targets for rejuvenating a given low-resolution (LR) input image to its high-resolution (HR) counterpart.
The task boasts a wide range of application, including but not limited to medical~\cite{medical1, medical2}, satellite~\cite{satellite1, satellite2} image processing, military surveillance~\cite{military}, and automotive industry, giving rise to a great amount of attention from the computer vision community.
Together with the interest from the community and the rapid development of deep neural networks (DNNs), SR has recently witnessed an unprecedented breakthrough in performance.

Recent SR works have enjoyed outstanding performance from increasing the network size~\cite{dong2014image,kim2016accurate,bee2017enhanced} and employing more complex network structures, such as residual block and spatial attention~\cite{zhang2018residual,zhang2018image,bee2017enhanced}.
However, a network with the larger size and complexity inherently demands heavier computing resources.
The deployment of such resource-hungry systems becomes challenging especially in practical applications that often have limited resources available.

Quantization has gained the popularity as one of the promising methods to reduce the amount of required resources for DNNs.
Through discretizing the values of network weights~\cite{hou2018loss}, features~\cite{choi2018pact}, or gradients~\cite{zhou2016dorefa},
quantization has significantly reduced the computation loads while minimizing the accuracy loss, particularly on high-level vision tasks, such as classification.
SR has yet to benefit from the recent advances in quantization mainly because SR requires pixel-level accuracy~\cite{xin2020binarized,ma2019efficient} and most SR methods have developed ad-hoc network structures~\cite{Li2020PAMSQS}, such as the absence of batch normalization (BN)~\cite{bee2017enhanced}.
\begin{figure}[t]
\begin{center}
\includegraphics[clip,trim=1.5cm 5cm 8cm 0cm, width=0.5\textwidth]{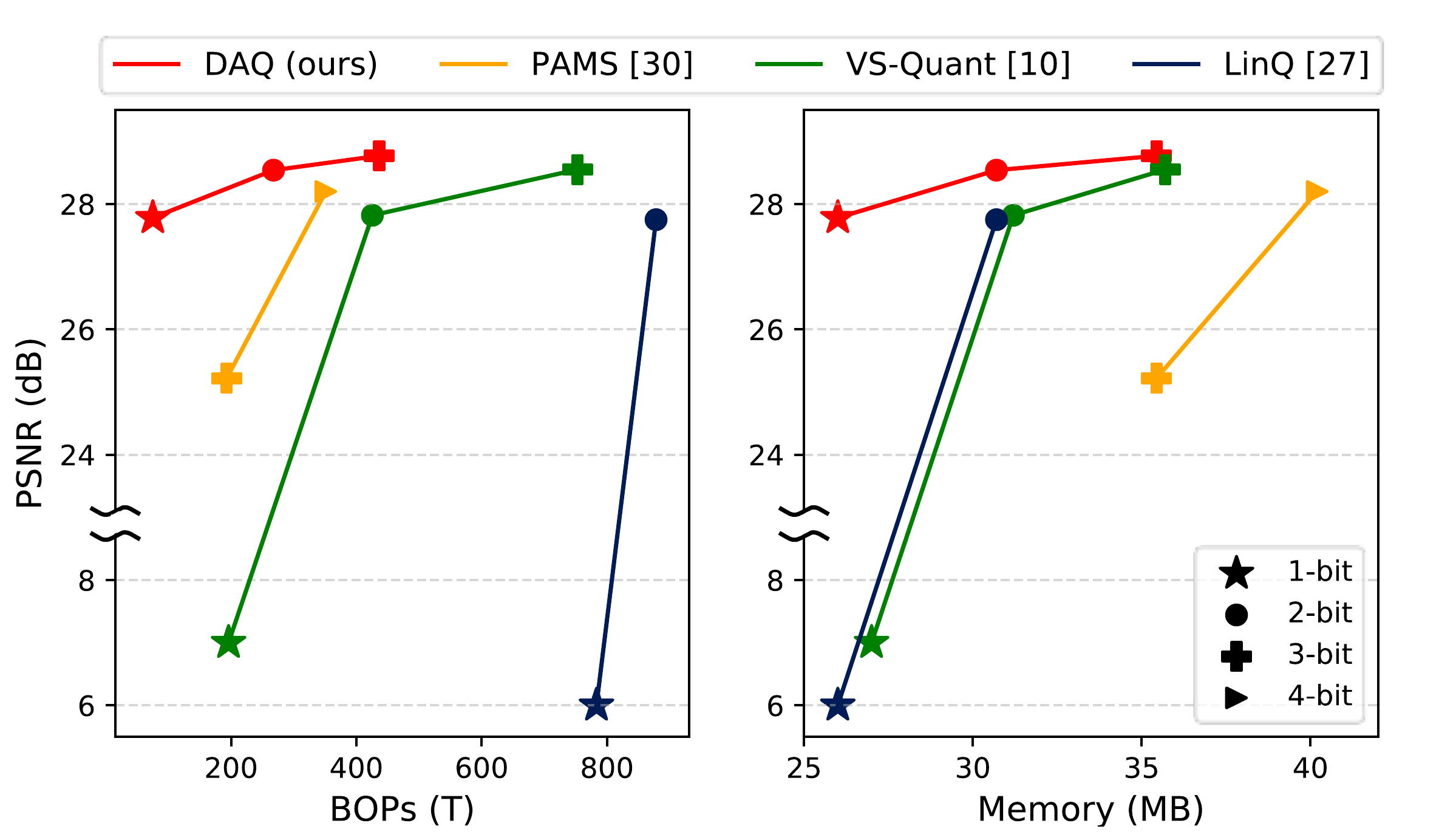}
\end{center}
\caption{
Quantizing EDSR($\times$4)~\cite{bee2017enhanced} on Set14.
The proposed quantization method (\textcolor{red}{{\textbf{DAQ}}}) achieves higher PSNR with less resource consumption (BOPs and memory).
}
\vspace{-0.5cm}
\label{fig:resrc-performance}
\end{figure}
\begin{figure}[t]      
\begin{center}
\setlength{\tabcolsep}{0.02cm}
        \setlength{\columnwidth}{2.81cm}
        \hspace*{-\tabcolsep}\begin{tabular}{cc}
        &
        \includegraphics[clip,trim=5.3cm 5cm 6cm 4cm, width=0.45\textwidth]{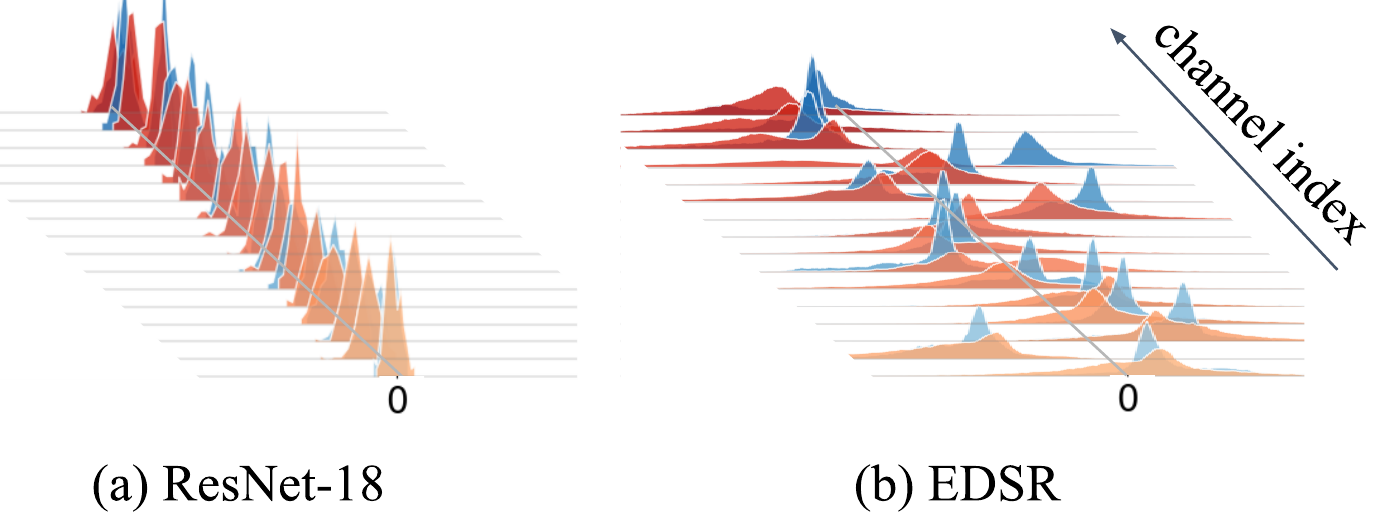}
        \end{tabular}
\vspace{-0.5cm}
\end{center}
  \caption{
  Channel-wise feature map distributions of two distinct images (\textcolor{red}{red} and \textcolor{blue}{blue}) in pre-trained (a) ResNet~\cite{he2016deep} on image classification task and (b) EDSR~\cite{bee2017enhanced} on image SR task. 
  In SR networks, channels present diverse non-zero distributions that also vary upon the input image.
  }
   \vspace{-0.7cm}
  \label{fig:observation}
\end{figure}
Few recent works have attempted to achieve quantization for SR via either adopting a learnable parameter for each binary convolution weight (BinarySR~\cite{ma2019efficient}); employing a bit accumulation mechanism to approximate the full-precision convolution (BAM~\cite{xin2020binarized}); and learning the layer-wise max scale value of weights and features (PAMS~\cite{Li2020PAMSQS}).
However, these methods have neglected the unique distribution characteristics of SR networks, which have domain-specific designs.

On the other hand, this work starts with an observation that the recent state-of-the-art SR networks have distinct distributions for each channel, compared to networks that are designed for semantic-level tasks, as illustrated in Figure~\ref{fig:observation}. 
Further, we observe that the channel distributions vary upon input images.

Upon the observations, we propose a \textbf{D}istribution-\textbf{A}ware \textbf{Q}uantization (DAQ) that performs an effective channel-wise quantization.
In particular, DAQ computes the distribution of each channel to determine the per-channel quantization transformation (or scaling) parameters that are adaptive to each input. 
The adaptive scaling approach facilitates effective clipping of outliers that dominate the quantization error.
Despite such compelling observation, per-channel quantization for feature map has yet to be deeply explored, due to the large computational overhead that could nullify the computation benefits of quantization~\cite{nagel2021white}.
We adjust the process of quantized convolution to significantly reduce the computational overhead, enabling DAQ to effectively \textit{and} efficiently quantize SR networks.

Overall, our contributions can be summarized as follows:
\begin{itemize}[noitemsep,nolistsep]
    \item To the best of our knowledge, we present the first feasible channel-wise quantization method for image super-resolution networks in ultra-low precision with marginal accuracy loss.
    \item Our scheme is directly applicable to existing SR networks without any architectural modification or specialized training scheme.
    \item Experimental results show that the proposed DAQ outperforms state-of-the-art SR quantization methods, yielding higher PSNR with less resource consumption.
    \item DAQ effectively reduces quantization error, achieving accurate quantization on pre-trained SR networks even without retraining.
\end{itemize}

\section{Related Work}
\label{sec:related_work}

\subsection{Image super-resolution}
\label{lightweight-SR}
Image super-resolution (SR) is a conventional computer vision problem that increases the resolution of an image while restoring its structural details.
Convolutional neural network (CNN) based approaches~\cite{bee2017enhanced,zhang2018residual,dong2014image,kim2016accurate} have achieved a great performance improvement in this field.
However, CNNs require heavy computational costs and memory footprints that greatly limit their applicability on resource-constrained devices.
To alleviate this problem, recent works~\cite{dong2016accelerating, shi2016realtime, hui2019lightweight, chu2019fast,guo2020hierarchical,kim2019fine} have focused on improving the model efficiency in terms of FLOPs, parameters, and run-time by designing novel lightweight network architectures.
However, these networks still consume a large amount of resources since the convolution layers of CNNs operate in 32-bit floating-point (FP32).

\subsection{Network quantization}
\label{network-quantization}
Network quantization is a promising research area that aims to map 32-bit floating-point (FP32) values of feature maps and weights to lower bits (or precision) values.
However, quantized networks usually suffer from severe accuracy degradation.
Many works tackle this problem by optimizing the uniform~\cite{courbariaux2015training, zhou2016dorefa, choi2018pact} or non-uniform~\cite{han2016deep, zhang2018lq, jung2019learning, esser2019learned} intervals between the quantized values.
This work focuses on quantization with the uniform interval that can be accelerated in the hardware with simple arithmetic pipelines.
Recently, research on specializing the precision of each part of the network~\cite{cai2020rethinking, lou2019autoq} and learning the quantization method without network retraining~\cite{krishnamoorthi2018quantizing, zhao2019improving, banner2019post, choukroun2019low, nagel2019data, cai2020zeroq} have been actively conducted.
However, these approaches focus on semantic-level tasks, such as image classification, which may be an easier task for network quantization than SR, which requires pixel-level accuracy~\cite{ma2019efficient,xin2020binarized}.

\begin{figure*}[t]      
\begin{center}
\includegraphics[clip,trim=0cm 0cm 0cm 0cm, width=1\textwidth]{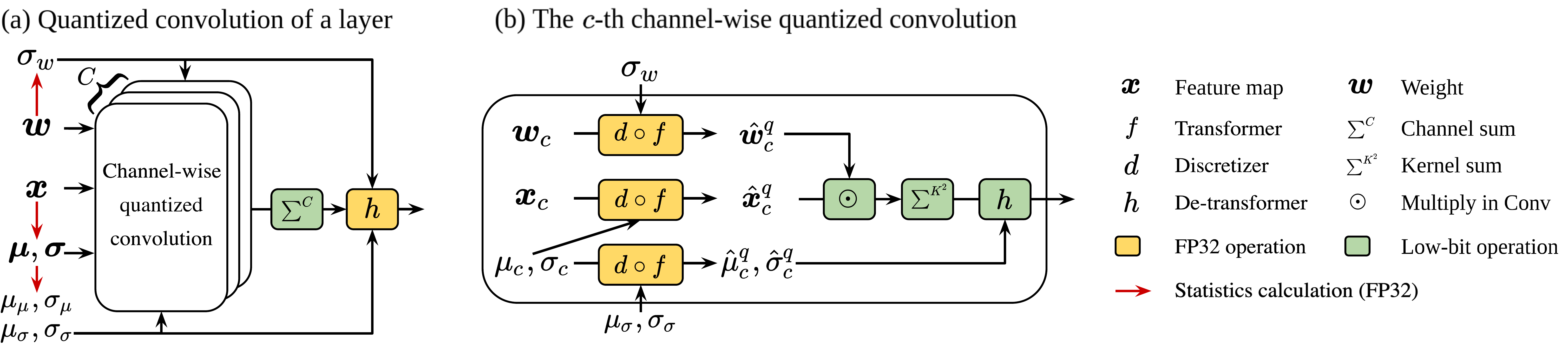}
\vspace{-0.8cm}
\end{center}
\caption{
Overview of our distribution-aware quantization (DAQ) process.
}
\vspace{-0.2cm}
\label{fig:overview}
\end{figure*}

To this end, few recent works have focused on network quantization specifically for SR.
BinarySR~\cite{ma2019efficient} presented the first attempt to quantize SR networks, however for weights only.
BAM~\cite{xin2020binarized} and BTM~\cite{jiang2021training} either modified the network architecture with multiple skip connections or utilized specialized convolution layers to quantize both weights and feature maps.
Recently, Hwang~\emph{et al.}~\cite{hwang2021layer} proposed an SR network compression method that jointly performs channel pruning and different-bit quantization for each layer.
The conventional approaches have neglected the first and last layers, but FQSR~\cite{Wang2021fully} quantized all the weights and feature maps of SR networks.
PAMS~\cite{Li2020PAMSQS} may be considered to be similar to our proposed method in that it learns quantization ranges with uniform intervals, however for each layer in contrast to our method that performs channel-wise quantization, which aligns well with the observation from Figure~\ref{fig:observation}.
Furthermore, the works mentioned above have proposed specialized training schemes such as knowledge distillation, self-supervised learning, and gradient update rules.
Although these custom network architectures and training schemes may reduce performance degradation to some extent, their ad-hoc techniques or specialized networks generally make it difficult to apply quantization methods to various networks.

On the other hand, we investigate a quantization range tailored to each channel in SR networks, while existing SR-based methods quantized all channels in a feature map with the same range.
Very few recent works have managed to achieve per-channel quantization, however for semantic-level tasks~\cite{wang2019learning}. 
To the best of our knowledge, our method is the first to achieve per-channel scaled quantization that manages to greatly reduce performance degradation. 

\section{Method}
\subsection{Motivation}
Network quantization for image super-resolution (SR) has suffered from substantial performance degradation that occurs due to quantization of feature map values~\cite{ma2019efficient}.
Most previous works~\cite{xin2020binarized, jiang2021training} tackle this problem by manually designing architectures and heuristic training schemes while their quantization functions are \textit{fixed} for a feature map. 
Our work started by investigating the characteristics of the feature maps in SR network.

The distribution of each channel in SR network (See Figure~\ref{fig:observation}\textcolor{red}{(b)}) has diverse values for its mean and variance.
Moreover, channel distributions are observed to be different for each input image.
This observation has led to our assumption that quantizing each channel in different functions may effectively reduce the quantization error in SR networks while it may not be as beneficial in semantic-level tasks.
To this end, we formulate a quantized convolution method as described in following sections.
\subsection{Distribution-Aware Quantization (DAQ)}
\label{sec:daq}
\subsubsection{Overview}
Quantization function is a sequence of processes that discretizes a FP32 tensor to lower precision.
General pipeline of quantization on neural networks are processed as following.
Input feature and weight are respectively scaled (or transformed) to integer range, and then discretized.
Then, the quantized convolution is performed with discretized input feature of integer values and discretized weight of integer values. 
Then the convolution output is descaled back via de-transformation, which is a costly procedure for channel-wise quantization. 
Our proposed method follows a general pipeline of quantization function, except that a standard de-transformation is replaced with our newly proposed efficient de-transformation scheme.
Overall, our DAQ convolution operates in a sequence of the following processes, as summarized in Figure~\ref{fig:overview}\textcolor{red}{(a)}:
\begin{itemize}
\vspace{-0.1cm}
\item Channel-wise quantized convolution (Figure~\ref{fig:overview}\textcolor{red}{(b)}). 
\begin{enumerate}
\item Per-channel transformation and discretization of input feature maps (Section~\ref{sec:feat_quant})
\item Transformation and discretization of convolution filter weights (Section~\ref{sec:feat_weight}).
\item Per-channel convolution over quantized weights and quantized input feature maps (dot product operation in a sliding window manner: multiplication of the two and kernel-wise sum).
\item Proposed computationally efficient per-channel de-transformation (Section~\ref{sec:de-transformer}).
\end{enumerate}
\vspace{-0.1cm}
\item Element-wise sum of outputs of the channel-wise convolution ($\sum^C$) in low-bit precision (Section~\ref{sec:de-transformer}).
\vspace{-0.1cm}
\item De-transformation of the convolution output to full precision FP32.
\vspace{-0.5cm}
\end{itemize}

\begin{figure*}[ht]      
\begin{center}
\includegraphics[clip,trim=0cm 5cm 0cm 0.5cm, width=1.0\textwidth]{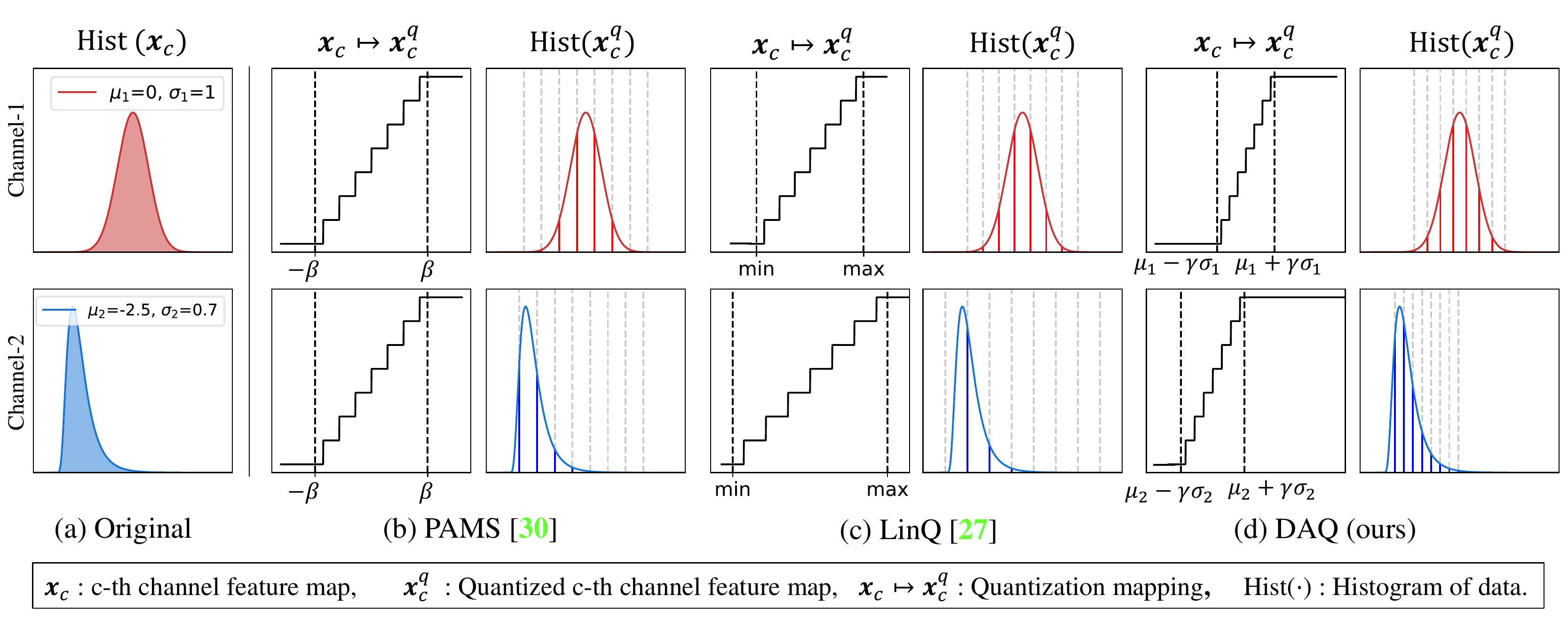}
\vspace{-1.2cm}
\end{center}
\caption{
Illustration of feature map quantization ranges with examples of two distribution from different channels (\textcolor{red}{red} and \textcolor{blue}{blue}).
(b) PAMS~\cite{Li2020PAMSQS} exploits learnable layer-wise fixed scale parameter $\beta$.
(c) channel-wise linear quantization (LinQ)~\cite{miyashita2016convolutional} 
utilizes the whole input tensor range as the quantization range, keeping the outliers that dominate quantization error.
(d) Our DAQ adaptively determines the channel-wise quantization range with $\mu$ and $\sigma$ that effectively clip outliers. $\gamma=2^{n-1}\cdot{s(n)}$ is the network hyperparameter.
}
\label{fig:method}
\vspace{-0.4cm}
\end{figure*}
\subsubsection{Quantization of feature map}
\label{sec:feat_quant}
As we discussed above, we define a quantization function specialized in each channel.
Formally, given a feature map $\boldsymbol{x} \in \mathbb{R}^{B\times C\times H\times W}$, we quantize $b$-th element of the mini-batch and $c$-th channel $\boldsymbol{x}_{b,c}\in \mathbb{R}^{H\times W}$, where $B, C, H,$ and $W$ denote the mini-batch size, number of channels, height and width of the feature map, respectively.
In the subsequent sections, we drop the subscript $b$ for notational simplicity.

\paragraph{Adaptive transformer}
Existing quantization works for SR~\cite{Li2020PAMSQS} often adopt linear scaling before discretizing the FP32 values to integer values.
The linear scaling might be a good transformer for tensor with a zero-mean and bell-shaped distribution, like the channel-wise feature of ResNet visualized in Figure~\ref{fig:observation}\textcolor{red}{(a)}.
However, the distribution of each channel, as shown in Figure~\ref{fig:observation}\textcolor{red}{(b)}, has a non-zero mean and even varies across different input images, suggesting that an adaptive transformer might be a better alternative for each channel distribution.
We define the transformer $f(\cdot)$ using a standardization function to normalize values of $\boldsymbol{x}_c$ before discretization.
$f(\cdot)$ is formally given by,
\begin{equation}
    \hat{\boldsymbol{x}}_c  \equiv f(\boldsymbol{x}_c)=\frac{\boldsymbol{x}_c-\mu_c}{\sigma_c \cdot s(n)},
\label{eq:transformer}
\end{equation}
where $\mu_c$ and $\sigma_c$, respectively, denote the average and the standard deviation of the $c$-th channel $\boldsymbol{x}_c$ and $s(n)$ determines an uniform interval between quantized values by the bit-width $n$.
Table~\ref{tab:method-step_size} presents the actual values of $s(n)$ which represent the optimal quantization for Gaussian distribution.
We interchangeably refer to statistics $\mu_c$ and $\sigma_c$ as quantization transformation parameters as they are used to perform transformation during quantization process.

\paragraph{Adaptive discretizer}
In general, a discretizer clips the values of $\hat{\boldsymbol{x}}_c$ to a range given by the quantization bit-width $n$, followed by the mapping of the values to integers.
Our adaptive discretizer is formulated as follows:
\begin{equation}
    \hat{\boldsymbol{x}}^q_c \equiv d(\hat{\boldsymbol{x}}_c)= \lfloor g(\hat{\boldsymbol{x}}_c) \rceil ,
\label{eq:discretizer}
\end{equation}
where $g(\cdot)$ is the clamp function of a range $(-2^{n-1}+\alpha,2^{n-1}+\alpha]$ and $\lfloor\cdot\rceil$ rounds up a value to the nearest higher integer.
The shifting parameter $\alpha$ is a function of $s(n)$, $\mu_c$, and, $\sigma_c$ formally given by,
\begin{equation}
\alpha= 
      \left\{ 
          \begin{matrix}
              \mathrm{max}(2^{n-1}-\frac{\mu_c}{\sigma_c\cdot s(n)} -1,0) & \mathrm{after\;ReLU} \\
              0  & \mathrm{otherwise,}
          \end{matrix}
      \right.
\label{eq:discretization-range}
\end{equation}
allowing the minimum quantized value (of original scale) to be zero after ReLU function.
$d(\cdot)$ is distribution-dependent due to its dependency on channel statistics, $\mu_c$ and $\sigma_c$.

Figure~\ref{fig:method} illustrates the effectiveness of our distribution-aware quantization (DAQ).
Our method reduces quantization error by a large margin for diverse channel distributions.
We use $\boldsymbol{x}^q_c  \equiv f^{-1}(\hat{\boldsymbol{x}}^q_c)$ for visualization, where $\hat{\boldsymbol{x}}^q_c$ denotes low-bit quantized integer and $f$ is the transform function of each method.

\begin{table}[t]
\caption{Step sizes $s(n)$ with respect to bit-width $n$. The sizes are the optimal values for uniform quantization of Gaussian distribution~\cite{lin2016fixed}.}
\centering
\vspace{-0.2cm}
\scalebox{0.95}{
\begin{tabular}[t]{l|c|c|c|c|c}
\hline
Bit-width $n$ & 1 & 2 & 3 & 4 & 8\\
\hline
Step size $s(n)$ & 1.596 & 0.996 & 0.586 & 0.335 & 0.031 \\
\hline
\end{tabular}}
\vspace{-0.2cm}
\label{tab:method-step_size}
\end{table}

\subsubsection{Quantization of weight}
\label{sec:feat_weight}
Weight distributions over filters are empirically observed to have less degree of variance than feature maps in SR networks.
Hence, a single quantization function is applied to the weight of each convolution $\boldsymbol{w}\in \mathbb{R}^{{C}\times {C}_{out}\times K\times K}$, where $C$ and $C_{out}$ denote the number of input and output channels, respectively, and $K$ is the kernel size of the convolution filter.
The distribution of weight values are observed to be bell-shaped with zero-mean, and is independent of the input image.
Thus, for quantizing the weight $\boldsymbol{w}$,
we assume the mean of $\boldsymbol{w}$ to be 0, which simplifies a transformer for weight quantization $f(\boldsymbol{w})$ as $f(\boldsymbol{w})=\frac{\boldsymbol{w}}{\sigma_w\cdot s(n)}$, where $\sigma_w$ denotes the standard deviation of $\boldsymbol{w}$.
The shifting parameter $\alpha$ of the weight discretizer is set to 0 since the weight values are not zeroed out by activation.

\subsubsection{Adaptive de-transformer}
\label{sec:de-transformer}
\paragraph{Standard per-channel de-transformer}
Once per-channel quantized convolution response is obtained, normally one would de-scale (or de-transform) each per-channel convolution output back to its full-precision before the summation of channel-wise convolution outputs along the channel dimension.
However, de-transformation of each channel-wise convolution output to full precision (FP32) using quantization transform parameters (in this case, channel-wise statistics $\mu_c$ and $\sigma_c$) will result in a large computational overhead of the overall quantized convolution process, due to the FP32 summation of de-transformed convolution responses along the channel dimension.

\paragraph{Efficient per-channel de-transformer}
Recently, Dai \textit{et al.}~\cite{dai2021vs} reduces the overhead by decomposing scaling factors into channel-specific quantized values and a global FP32 value for the low-bit addition in image classification networks.
Similarly, our solution achieves efficient implementation but with the distribution awareness.
We propose to perform quantization on statistics $\boldsymbol{\mu}$ and $\boldsymbol{\sigma}$ to obtain quantized statistics, allowing us to perform the summation of the channel-wise convolution responses in low-bit precision.
Formally, recall $\mu_c$ and $\sigma_c$, the mean and the standard deviation of the $c$-th channel in Equation~\eqref{eq:transformer}, as the $c$-th elements of $\boldsymbol{\mu}\in \mathbb{R}^{C}$ and $\boldsymbol{\sigma}\in \mathbb{R}^{C}$.
We \textbf{q}uantize the \textbf{q}uantization transform parameters, thus named QQ.
The quantized transform parameters using Equation~\eqref{eq:transformer} and~\eqref{eq:discretizer} approximate their FP32 versions:
\vspace{-1em}
\begin{equation}
    \mu_c \approx \mu^q_c \equiv \sigma_{\mu} \cdot s(m) \cdot \hat{\mu}^q_c + \mu_{\mu},
\label{eq:qq}
\end{equation}
\vspace{-1.5em}
\begin{equation}
    \sigma_c \approx \sigma^q_c \equiv \sigma_{\sigma} \cdot s(m)  \cdot \hat{\sigma}^q_c + \mu_{\sigma},
\label{eq:qq2}
\end{equation}
where $m$ is the bit-width of QQ, and $\sigma_*$ and $\mu_*$, $* \in \{\mu,\sigma\}$, denote the mean and the standard deviation of $\boldsymbol{\mu}$ and $\boldsymbol{\sigma}$.
Note that $\sigma_*$ and $\mu_*$, $* \in \{\mu,\sigma\}$, is FP32 values which are independent to channels while $\hat{\sigma}^q_c$ and $\hat{\mu}^q_c$ are low-bit integers specific to the channels.
Using the quantizing quantization parameters, our quantized channels become,
\begin{equation}
\begin{array}{rl}
    \boldsymbol{x}^q_c = \sigma_c \cdot s(n) \cdot \hat{\boldsymbol{x}}^q_c  + \mu_c 
    \\[0.5em] \approx \sigma^q_c \cdot s(n) \cdot \hat{\boldsymbol{x}}^q_c  + \mu^q_c.
\end{array}
\label{eq:qq3}
\end{equation}

As a result, we can place the de-transformation process after the summation of the channel-wise convolution and eliminate the potential possibility of a computation overhead that could have occurred from channel-wise quantized convolution (please refer to Section~\ref{sec:qq} and the supplementary document).

\subsection{Hardware Implementation}
\label{sec:hardware}
\subsubsection{Computation of per-channel detransformer}
\label{sec:qq}
As discussed in Section~\ref{sec:de-transformer}, the computational overhead from the standard per-channel de-transformer is largely reduced by our efficient adaptive de-transformation process.
The approximation by Equation~\eqref{eq:qq3} replaces a large proportion of high-precision operations with low-precision operations by modifying arithmetic pipelines.
For instance, the sum of quantized channels ($\sum_c \boldsymbol{x}^q_c$) should be operated in FP32 ($\sum_c (\sigma_c \cdot s(n) \cdot \hat{\boldsymbol{x}}^q_c + \mu_c)$) due to the channel-specific mean and standard deviation of FP32 values.
However, the sum of the approximated quantized channels ($\sum_c (\sigma^q_c \cdot s(n) \cdot \hat{\boldsymbol{x}}^q_c + \mu^q_c)$) can expand the low-bit channel sum ($\sigma_\sigma \cdot s(n)\cdot s(m) \sum_c \hat{\sigma}^q_c \cdot \hat{\boldsymbol{x}}^q_c + \mu_\sigma \cdot s(m) \sum_c \hat{\boldsymbol{x}}^q_c + \sigma_{\mu} \cdot s(m) \sum_c \hat{\mu}^q_c + \mu_{\mu}$).

\subsubsection{FP32 Arithmetic Computation}
The proposed method adaptively quantizes the value of interest with respect to its distribution by using standardization function as the tranformer.
The computation complexity of Equation~\eqref{eq:transformer} can be specified into the complexity for calculating the mean and the standard deviation, and for the transformer function itself, of which are both $O(n)$.
Specifically, the overall complexity of Equation~\eqref{eq:transformer} is $C(5HW+3)$ for the feature map and $3K^2 C_{in} C_{out}$ for the weight.
Weight standardization is independent of input images and thus occurs only once before the inference, which allows us to measure its computational complexity only once.
On the other hand, the complexity of feature standardization is calculated per test image.

\subsection{Training}
\label{sec:training}
Recently, training quantized networks (or \textit{quantization-aware training}) is a well-known approach for accuracy recover. 
The back-propagation in quantized networks calculates the gradient of loss $\ell$ with respect to $\boldsymbol{w}^q$. 
Quantization function is generally not differentiable, and thus not possible to directly apply back-propagation.
Consequently, we adopt the straight-through estimator~\cite{zhou2016dorefa} to approximate the gradients calculation. 
We approximate the partial gradient $\frac{\partial \boldsymbol{w}^q}{\partial{\boldsymbol{w}}}$ with an identity mapping ($\frac{\partial \boldsymbol{w}^q}{\partial{\boldsymbol{w}}} \approx 1$).
Accordingly, $\frac{\partial \ell}{\partial {\boldsymbol{w}^q}}$ can be approximated by 
\begin{equation}
    \frac{\partial \ell}{\partial \boldsymbol{w}^q}=\frac{\partial \ell}{\partial \boldsymbol{w}} \frac{\partial \boldsymbol{w}}{\partial \boldsymbol{w}^q} \approx \frac{\partial \ell}{\partial \boldsymbol{w}}.
\label{eq:STE}
\end{equation}

\section{Experiments}
\label{sec:experiments}

\begin{table*}[t]
\caption{Comparisons of existing quantization methods on EDSR~\cite{bee2017enhanced}, RDN~\cite{zhang2018residual} and SRResNet~\cite{ledig2017photo} of scale 4. 
}
\vspace{-0.2cm}
\centering
\scalebox{0.94}{
\begin{tabular}{l cc rrrc cccc}
\toprule
 \multirow{2}{*}{Method} &\multicolumn{2}{c}{Precision} & \multirow{2}{*}{BOPs} & \multirow{2}{*}{Energy } & \multirow{2}{*}{Memory} & \multirow{2}{*}{Parameters} & \multicolumn{4}{c}{PSNR (dB)} \vspace{-0.1cm}\\
\cmidrule(lr){2-3}  \cmidrule(lr){8-11}
\multicolumn{1}{c}{}& w& a& & & & & Set5 & Set14 & B100 & Urban100 \\
\midrule
EDSR & 32 & 32 &  10019.3 T & 22504.3 $m$J & 172.3 MB & 43.1M & 32.46 & 28.77 & 27.69 & 26.54 \\
EDSR-PAMS~\cite{Li2020PAMSQS} & 4 & 4 & 351.3 T & 366.7 $m$J& 40.2 MB & 43.1M & 31.59 & 28.20 & 27.32 & 25.32 \\
EDSR-LinQ~\cite{krishnamoorthi2018quantizing} &2&2 &878.4 T &2616.6 $m$J &30.7 MB &43.1M & 31.08&27.75&27.05&24.45\\
EDSR-VS-Quant~\cite{dai2021vs}&2&2 &425.6 T &555.7 $m$J &31.0 MB & 43.1M & 31.10&27.82&27.11&24.94 \\
\rowcolor{orange!25}
EDSR-DAQ (ours) & 2&2& 267.8 T & 328.6 $m$J & 30.7 MB & 43.1M & 32.05 & 28.54 & 27.50 & 25.97 \\
\midrule
RDN & 32 & 32 & 5636.0 T& 12659.0 $m$J & 89.2 MB &22.3M &32.24 &28.67 & 27.63 & 26.29\\
RDN-PAMS~\cite{Li2020PAMSQS} & 4 & 4 & 204.8 T& 222.3 $m$J& 12.3 MB& 22.3M & 30.44&27.54&26.87&24.52\\
RDN-LinQ~\cite{krishnamoorthi2018quantizing} &2&2 &474.5 T &1044.3 $m$J &6.9 MB &22.3M & 30.90&27.73&27.05&24.65\\
RDN-VS-Quant~\cite{dai2021vs} &2&2 & 239.1 T &311.6 $m$J&6.9 MB&22.3M&31.16&27.89&27.12&24.83 \\
\rowcolor{orange!25}
RDN-DAQ (ours) & 2&2& 168.8 T& 220.1 $m$J& 6.9 MB& 22.3M& 31.61& 28.21& 27.31&25.52\\
\midrule
SRResNet &32&32& 324.6 T& 728.2 $m$J& 6.1 MB & 1.6M & 31.94& 28.43 & 27.46 & 25.71\\
SRResNet-BinarySR~\cite{ma2019efficient} &1&32& 38.9 T& 58.3 $m$J& 1.5 MB & 1.6M& 30.16& 27.19&26.66 &24.24\\
SRResNet-LinQ~\cite{krishnamoorthi2018quantizing}&2&2 & 37.8 T &58.8 $m$J &1.7 MB & 1.6M & 31.44&	28.03&	27.21&	25.05 \\
SRResNet-VS-Quant~\cite{dai2021vs}&2&2 &22.5 T &41.8 $m$J &1.7 MB & 1.6M &31.24&	27.95	&27.15&	24.89 \\
\rowcolor{orange!25}
SRResNet-DAQ (ours) &2&2& 19.1 T& 38.5 $m$J &1.7 MB& 1.6M&  31.67&28.26&27.32&25.39\\
\bottomrule
\end{tabular}}
\label{tab:exp-main}
\end{table*}

\subsection{Experimental setup}
\paragraph{Models and datasets}
To verify the effectiveness and the generalizability of the proposed quantization method, we conduct experiments on widely used SR networks, including EDSR~\cite{bee2017enhanced}, RDN~\cite{zhang2018residual} and SRResNet~\cite{ledig2017photo}.
We use pre-trained models publicly available at codes\footnote{https://github.com/thstkdgus35/EDSR-PyTorch}.
We retrain the quantized models on DIV2K dataset~\cite{agustsson2017ntire}, which consists of 800 2K-resolution images.
We finetune the DAQ-applied EDSR model using the training settings from~\cite{bee2017enhanced}, where a batch size is set to be 4, a learning rate is $10^{-4}$, and the number of updates is $3\times 10^{5}$.
DAQ-applied RDN is finetuned using $3\times 10^{4}$ updates with batch size 16, and initial learning rate $10^{-4}$.
SRResNet-DAQ is finetuned with an initial learning rate of $10^{-5}$ for $6\times 10^{5}$ iterations while we follow the other training settings from~\cite{ledig2017photo}.
We evaluate the quantized models on four standard benchmark datasets: Set5, Set14, B100, and Urban100.

\paragraph{Implementation details}
Our proposed quantization method is directly applied to existing SR models without modifying the network architecture.
Along with the majority of quantization works, we quantize weights and feature maps in all layers except the first convolution and the last reconstruction modules.
We set $m\Equal4$ for quantizing quantization parameters.

\paragraph{BOPs} 
The number of floating-point operations (FLOPs) is a conventional measure for the computational complexity of full-precision models. 
However, a low-precision model often consists of operations with inputs of different low bit-widths.
To take the low-precision operations into account, we weight the number of bit operations by multiplying $n\cdot m$, where $n$ and $m$ denote the bit-widths of two different inputs, respectively for each operation. The weighted number of bit operations is referred to as BOPs.
The reported BOPs are measured to generate a FHD image (1920$\times$1080).

\paragraph{Energy}
A quantized network boasts energy efficiency as lower-precision operations consume less energy.
We adopt the approximation of energy consumption introduced in~\cite{dally2015high, horowitz2014computing} and compute it for generating a FHD image.

\paragraph{Memory}
Another benefit of network quantization is the reduction in memory requirement for model storage.
The $n$-bit weights require $\frac{n}{32}$ amount of memory of FP32 weights.

\subsection{Comparisons with SotA methods}
\paragraph{SR quantization methods}
To demonstrate the effectiveness of our channel-wise quantization paradigm, we compare our method with PAMS~\cite{Li2020PAMSQS} and BinarySR~\cite{ma2019efficient}, which adopt per-layer quantization ranges.
Without architecture changes of existing networks, Table~\ref{tab:exp-main} presents outstanding performances of our method (DAQ) on most measures and all benchmarks.
Specifically, RDN-DAQ achieves $1$ dB higher PSNR than RDN-PAMS on Urban100 with less resource consumption of BOPs, Energy, and Memory.
SRResNet-DAQ requires only 50 \% BOPs of SRResNet-BinarySR of which weights are 1-bit while outperforming $1.51$ dB on Set5.


\begin{figure*}[ht]
\begin{center}\centering
\setlength{\tabcolsep}{0.05cm}
\hspace*{-\tabcolsep}
\begin{tabular}{ccccc}
\includegraphics[width=0.18\linewidth]{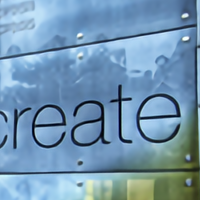} & 
\includegraphics[width=0.18\linewidth]{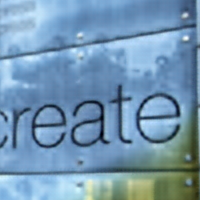} &
\includegraphics[width=0.18\linewidth]{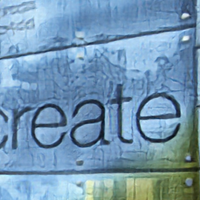} & 
\includegraphics[width=0.18\linewidth]{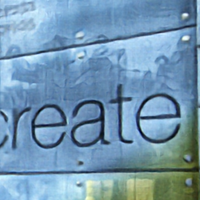} &
\includegraphics[width=0.18\linewidth]{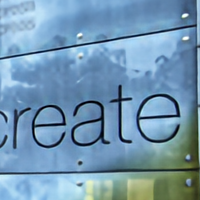} \vspace{-0.1cm}
\\
\footnotesize EDSR & \footnotesize EDSR-PAMS (w4a4) &\footnotesize EDSR-LinQ (w2a2) &\footnotesize EDSR-VS-Quant (w2a2) &\footnotesize EDSR-DAQ (w2a2) \vspace{-0.1cm} \\
\footnotesize (22.08 dB / 193.3 T) & \footnotesize (20.95 dB / 6.78 T) &\footnotesize  (20.64 dB / 16.94 T) &\footnotesize  (21.29 dB / 8.21 T) &\footnotesize (\textbf{21.84} dB / \textbf{5.17} T) 
\vspace{+0.2cm} \\
\includegraphics[width=0.18\linewidth]{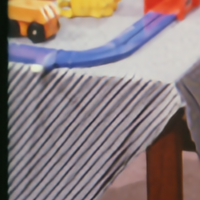} & 
\includegraphics[width=0.18\linewidth]{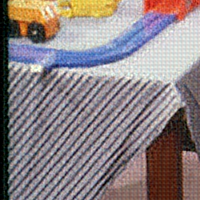} &
\includegraphics[width=0.18\linewidth]{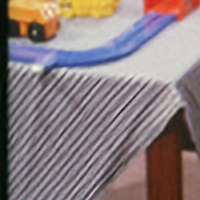} & 
\includegraphics[width=0.18\linewidth]{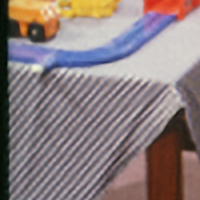} &
\includegraphics[width=0.18\linewidth]{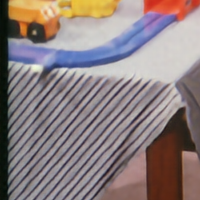}  \vspace{-0.1cm}
\\
\footnotesize RDN & \footnotesize RDN-PAMS (w4a4) &\footnotesize RDN-LinQ (w2a2) &\footnotesize RDN-VS-Quant (w2a2) &\footnotesize RDN-DAQ (w2a2) \vspace{-0.1cm} \\
\footnotesize (27.70 dB / 108.72 T) & \footnotesize (22.23 dB / 3.95 T) &\footnotesize (24.90 dB / 9.15 T) &\footnotesize (25.01 dB / 4.61 T) &\footnotesize (\textbf{26.85} dB / \textbf{3.26} T)
\vspace{+0.2cm} \\
\includegraphics[width=0.18\linewidth]{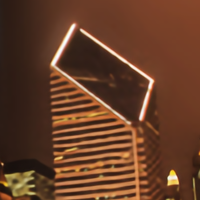} & 
\includegraphics[width=0.18\linewidth]{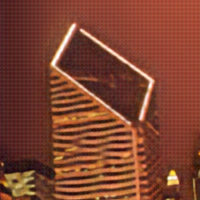} &
\includegraphics[width=0.18\linewidth]{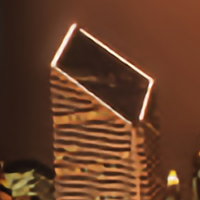} & 
\includegraphics[width=0.18\linewidth]{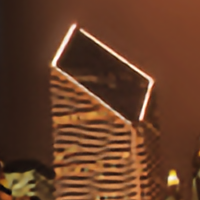} &
\includegraphics[width=0.18\linewidth]{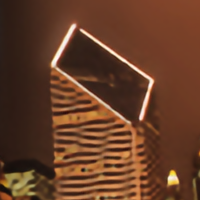} \vspace{-0.1cm}
\\
\footnotesize SRResNet & \footnotesize SRResNet-BinarySR (w1a32) &\footnotesize SRResNet-LinQ (w2a2) &\footnotesize SRResNet-VS-Quant (w2a2) &\footnotesize SRResNet-DAQ (w2a2) \vspace{-0.1cm} \\
\footnotesize (25.22 dB / 6.26 T) & \footnotesize (22.19 dB / 0.75 T) &\footnotesize (21.31 dB / 0.73 T) &\footnotesize (22.27 dB / 0.43 T) &\footnotesize (\textbf{23.23} dB / \textbf{0.37} T)  
\end{tabular}
\captionof{figure}{
Qualitative evaluation of existing quantization methods on EDSR~\cite{bee2017enhanced}, RDN~\cite{zhang2018residual} and SRResNet~\cite{ledig2017photo} of scale 4.
Evaluation is done on `barbara' of Set14 and `img060' and `img097' of Urban100. Quantitative measures are also denoted (PSNR / BOPs).
}\vspace{-1cm}
\label{fig:qualitative}
\end{center}
\end{figure*}

On the other hand, BTM~\cite{jiang2021training} and BAM~\cite{xin2020binarized} employ specialized architectures or training schemes for low-precision SR networks.
Our proposed quantization method is orthogonal to these techniques, providing additional performance gains, as shown in the supplementary document.

\paragraph{Comparison with per-channel quantization methods}
To show the importance of distribution awareness in per-channel quantization function as proposed in this work, we compare DAQ with the SR models quantized by other per-channel quantization methods LinQ~\cite{krishnamoorthi2018quantizing} and VS-Quant~\cite{dai2021vs}, which are however tailed for classification and do not consider per-channel distribution in contrast to DAQ.
Table~\ref{tab:exp-main} shows that EDSR-DAQ outperforms EDSR-LinQ and EDSR-VS-Quant over than 1 dB on Urban100 while consuming less resources, validating the benefits of formulating the channel-wise quantization function with distribution statistics.

\begin{table*}[ht]
\caption{ Comparison of quantization methods \textit{without retraining} on pre-trained EDSR~\cite{bee2017enhanced} of scale 4.
}
\vspace{-0.2cm}
\centering
\scalebox{0.83}{
\begin{tabular}{l cc ccc cccc}
\toprule
\multirow{2}{*}{Method} &\multicolumn{2}{c}{Precision} & \multirow{2}{*}{ BOPs} & \multirow{2}{*}{ Energy} & \multirow{2}{*}{Memory}& \multicolumn{4}{c}{PSNR (dB)} \vspace{-0.1cm}\\
\cmidrule(lr){2-3}  \cmidrule(lr){7-10}  
\multicolumn{1}{c}{}& w& a& && &Set5 & Set14 & B100 & Urban100 \\
\midrule
EDSR& 32 & 32 & 10019.3 T& 22504.3 $m$J&172.3 MB&32.46 & 28.77 & 27.69 & 26.54\\
EDSR-LinQ~\cite{krishnamoorthi2018quantizing} & 2 & 2  & 878.4 T& 2616.6 $m$J& 30.7 MB&30.26&	26.78&	26.44&	23.14 \\
EDSR-DFQ~\cite{nagel2019data}  & 4 & 4  & 351.3 T& 366.7 $m$J&40.2 MB& 31.20&	27.77&26.29&24.18 \\ 
\rowcolor{orange!25}
EDSR-DAQ  & 2 & 2  & 267.8 T& 328.6 $m$J&30.7 MB& 31.42&	28.12&	27.23&	25.56   \\ 
\bottomrule
\end{tabular}}
\vspace{-0.2cm}
\label{tab:exp-trainingfree}
\end{table*}
\paragraph{Qualitative comparisons}
Figure~\ref{fig:qualitative} presents the output images from SotA models compared in Table~\ref{tab:exp-main}. 
Our method generates a visually clean output image, while RDN-PAMS~\cite{Li2020PAMSQS} and SRResNet-BinarySR~\cite{ma2019efficient} suffer from a checkerboard artifact.
Interestingly, EDSR-LinQ generates a sharper image than EDSR-PAMS with lower PSNR scores.
The result of SRResNet-DAQ has a more realistic structure (at the bottom of the image).

\paragraph{Comparison with methods without retraining}
To further demonstrate the effectiveness of DAQ in dynamically reducing the quantization error, we evaluate DAQ and other quantization methods without fine-tuning, as shown in Table~\ref{tab:exp-trainingfree}.
DAQ is shown to greatly reduce the quantization error to a large extent compared to other methods. 
EDSR-LinQ calculates min/max values for each channel-wise tensor during inference, which is the same process in the previous section.
DFQ~\cite{nagel2019data} controls weight ranges and biases of the pre-trained models during inference, proposed for image classification.
EDSR-DFQ is a re-implemented version of DFQ on EDSR, based on the publicly available codes.
EDSR-DAQ outperforms the compared methods over 1.4 dB with less computation costs. 
The results corroborate our observations from channel distributions (Figure~\ref{fig:observation}) and assumptions that the per-channel and distribution-aware quantization is effective in SR networks.

\begin{table}[ht]
\caption{Ablation study of the proposed method on feature map quantization. 
2-bit quantization is evaluated on EDSR.
}
\vspace{-0.15cm}
\centering
\scalebox{0.85}{
\begin{tabular}{ccc rr}
\toprule
\multirow{2}{*}{Granularity}  & Distribution& \multirow{2}{*}{QQ} & \multirow{2}{*}{BOPs} &PSNR \\
 &aware scaling& & & (Urban100) \\
\midrule
Layer & \xm & \xm & 83.5 T&  6.99 dB\\
Layer & \cm & \xm & 90.1 T&  7.17 dB\\
\midrule
Channel &  \xm & \xm & 878.4 T& 7.26 dB\\
Channel &  \cm & \xm & 889.3 T& 25.98 dB\\
\rowcolor{orange!25}
Channel &  \cm & \cm & \textbf{267.8 T}& \textbf{25.97 dB}\\
\bottomrule
\end{tabular}
}
\label{tab:exp-ablation-act}
\vspace{-0.2cm}
\end{table}

\begin{table}[ht]
\caption{ Ablation study of the proposed method on weight quantization. 
Evaluation done on EDSR.
} 
\vspace{-0.15cm}
\centering
\scalebox{0.85}{
	\begin{tabular}{clccc}
		\toprule
		Distribution \vspace{-0.1cm}& \multirow{2}{*}{Granularity} & \multicolumn{3}{c}{Precision}\\ 
		\cmidrule(lr){3-5} 
		aware scaling & & 4-bit & 2-bit & 1-bit \\
		\midrule \rowcolor{orange!25}
		\cm & Layer &26.48&26.07&25.07\\ 
		 & Output channel &26.50&26.15&24.94\\
		 & Input channel  &26.47&26.10&25.09\\
		 & Kernel         &26.45&26.12&25.31\\
		\midrule
		\xm & Layer &26.20&7.16&7.17\\
		& Output channel &26.51&7.11&7.17\\
		& Input channel  &26.50&23.84&7.17 \\
		& Kernel         &26.54&26.47&17.73\\
		\bottomrule
	\end{tabular}
}
\vspace{-0.3cm}
\label{tab:exp-ablation-weight}
\end{table}

\begin{table}[ht]
\caption{Ablation study on various distribution assumptions.
}
\vspace{-0.15cm}
\centering
\scalebox{0.85}{
    \begin{tabular}{ccccc}
        \toprule
         EDSR~($\times$4) & Gaussian & Uniform & Laplacian & Gamma \\
        \midrule
        4-bit& 26.51 & 24.52 & 26.21 & 26.25 \\
        1-bit& 24.20 & 23.64 & 23.82 & 23.71 \\
        \bottomrule
    \end{tabular}
}
\vspace{-0.3cm}
\label{tab:exp-ablation-distribution}
\end{table}
\subsection{Ablation study}
\paragraph{Feature map quantization}
To verify the efficacy of our quantization scheme in practical SR networks, we do ablation on each of our contribution.
In Table~\ref{tab:exp-ablation-act}, granularity represents the tensor to be quantized, where ``Layer'' quantizes the feature map with a single quantization range and ``Channel'' quantizes each input channel with a specialized range.
The non-distribution-aware scaling uses min/max values to normalize distributions.
As shown in Table~\ref{tab:exp-ablation-act}, channel-wise and distribution-aware property are both important for accurate low-bit quantization, but suffer from heavy computational resources. 
Utilizing quantization function for the quantization transform parameters (QQ) significantly reduces BOPs, while maintaining a high PSNR.

\paragraph{Weight quantization}
Though our main contributions in the method are focused on feature map quantization, we also evaluate the effectiveness of the proposed weight quantization in Table~\ref{tab:exp-ablation-weight}.
Distribution awareness in quantization plays an key role in maintaining the performance, especially in lower-precisions.
Layer-wise quantization presents a better trade-off between computational efficiency and performance among different extents of granularity. 
\paragraph{Distribution assumption}
Scaled step size $s(n)$ in quantization function is a pre-determined hyper-parameter. 
Many quantization works~\cite{cai2017deep, dong2019hawqv2} select different scaled step size for different bit-widths 
assuming a certain distribution of the tensor to be quantized.
We follow the Gaussian assumption in the main experiments, while results on different distribution assumptions, including Gaussian, Uniform, Laplacian, and Gamma distribution are shown in Table~\ref{tab:exp-ablation-distribution}.
\section{Conclusion}
We propose an efficient yet effective quantization method for deep SR networks.
Based on the observation that each channel of SR networks has distinct distribution,
we introduce a channel-wise \textbf{D}istribution-\textbf{A}ware \textbf{Q}uantization (DAQ) method.
However, channel-wise quantization with standard de-transformation incurs a large computational overhead.
Instead, we utilize an efficient de-transformation by quantizing quantization parameters, reducing the computational costs of channel-wise quantization significantly.
Our proposed scheme reduces the quantization error to a large extent, achieving high performance even without quantization-tailored training.
Evaluations on various SR networks demonstrate the outstanding performance of DAQ, compared to other quantization methods, with similar amount of resource consumptions.

\paragraph{Acknowledgment}
\noindent 
This work was supported in part by IITP grant funded by the Korean government (MSIT) [NO.2021-0-01343, Artificial Intelligence Graduate School Program (Seoul National University)]

{\small
\bibliographystyle{ieee_fullname}
\bibliography{egbib}

\begin{thebibliography}{10}\itemsep=-1pt

\bibitem{agustsson2017ntire}
Eirikur Agustsson and Radu Timofte.
\newblock Ntire 2017 challenge on single image super-resolution: Dataset and
  study.
\newblock In {\em CVPRW}, 2017.

\bibitem{banner2019post}
Ron Banner, Yury Nahshan, and Daniel Soudry.
\newblock Post training 4-bit quantization of convolutional networks for
  rapid-deployment.
\newblock In {\em NIPS}, 2019.

\bibitem{satellite1}
K.~Vani C.~Heltin~Genitha.
\newblock Super resolution mapping of satellite images using hopfield neural
  networks.
\newblock {\em Recent Advances in Space Technology Services and Climate Change
  (RSTSCC)}, 2010.

\bibitem{cai2020zeroq}
Yaohui Cai, Zhewei Yao, Zhen Dong, Amir Gholami, Michael~W Mahoney, and Kurt
  Keutzer.
\newblock Zeroq: A novel zero shot quantization framework.
\newblock In {\em CVPR}, 2020.

\bibitem{cai2017deep}
Zhaowei Cai, Xiaodong He, Jian Sun, and Nuno Vasconcelos.
\newblock Deep learning with low precision by half-wave gaussian quantization.
\newblock In {\em CVPR}, 2017.

\bibitem{cai2020rethinking}
Zhaowei Cai and Nuno Vasconcelos.
\newblock Rethinking differentiable search for mixed-precision neural networks.
\newblock In {\em CVPR}, 2020.

\bibitem{choi2018pact}
Jungwook Choi, Zhuo Wang, Swagath Venkataramani, Pierce I-Jen Chuang,
  Vijayalakshmi Srinivasan, and Kailash Gopalakrishnan.
\newblock Pact: Parameterized clipping activation for quantized neural
  networks.
\newblock {\em arXiv preprint arXiv:1805.06085}, 2018.

\bibitem{choukroun2019low}
Yoni Choukroun, Eli Kravchik, Fan Yang, and Pavel Kisilev.
\newblock Low-bit quantization of neural networks for efficient inference.
\newblock In {\em ICCVW}, 2019.

\bibitem{chu2019fast}
Xiangxiang Chu, Bo Zhang, Hailong Ma, Ruijun Xu, Jixiang Li, and Qingyuan Li.
\newblock Fast, accurate and lightweight super-resolution with neural
  architecture search.
\newblock {\em arXiv preprint arXiv:1901.07261}, 2019.

\bibitem{courbariaux2015training}
Matthieu Courbariaux, Yoshua Bengio, and Jean-Pierre David.
\newblock Training deep neural networks with low precision multiplications.
\newblock In {\em ICLRW}, 2015.

\bibitem{dai2021vs}
Steve Dai, Rangha Venkatesan, Brian~Zimmer Mark~Ren, William Dally, and Brucek
  Khailany.
\newblock Vs-quant: Per-vector scaled quantization for accurate low-precision
  neural network inference.
\newblock In {\em MLSys}, 2021.

\bibitem{dally2015high}
William Dally.
\newblock High-performance hardware for machine learning.
\newblock In {\em NeurIPS Tutorial}, 2015.

\bibitem{dong2014image}
Chao Dong, Chen~Change Loy, Kaiming He, and Xiaoou Tang.
\newblock Image super-resolution using deep convolutional networks.
\newblock {\em TPAMI}, 2014.

\bibitem{dong2016accelerating}
Chao Dong, Chen~Change Loy, and Xiaoou Tang.
\newblock Accelerating the super-resolution convolutional neural network.
\newblock In {\em ECCV}. Springer, 2016.

\bibitem{dong2019hawqv2}
Zhen Dong, Zhewei Yao, Yaohui Cai, Daiyaan Arfeen, Amir Gholami, Michael~W.
  Mahoney, and Kurt Keutzer.
\newblock Hawq-v2: Hessian aware trace-weighted quantization of neural
  networks.
\newblock In {\em NeurIPS}, 2020.

\bibitem{esser2019learned}
Steven~K Esser, Jeffrey~L McKinstry, Deepika Bablani, Rathinakumar Appuswamy,
  and Dharmendra~S Modha.
\newblock Learned step size quantization.
\newblock {\em arXiv preprint arXiv:1902.08153}, 2019.

\bibitem{medical1}
Hayit Greenspan.
\newblock Super-resolution in medical imaging.
\newblock {\em The Computer Journal, Oxford University Press Oxford, UK}, 2009.

\bibitem{guo2020hierarchical}
Yong Guo, Yongsheng Luo, Zhenhao He, Jin Huang, and Jian Chen.
\newblock Hierarchical neural architecture search for single image
  super-resolution.
\newblock {\em arXiv preprint arXiv:2003.04619}, 2020.

\bibitem{han2016deep}
Song Han, Huizi Mao, and William~J. Dally.
\newblock Deep compression: Compressing deep neural networks with pruning,
  trained quantization and huffman coding.
\newblock In {\em ICLR}, 2016.

\bibitem{he2016deep}
Kaiming He, Xiangyu Zhang, Shaoqing Ren, and Jian Sun.
\newblock Deep residual learning for image recognition.
\newblock In {\em CVPR}, 2016.

\bibitem{horowitz2014computing}
Mark Horowitz.
\newblock 1.1 computing’s energy problem (and what we can do about it).
\newblock In {\em International Solid-State Circuits Conference Digest of
  Technical Papers (ISSCC)}, 2014.

\bibitem{hou2018loss}
Lu Hou and James~T. Kwok.
\newblock Loss-aware weight quantization of deep networks.
\newblock In {\em ICLR}, 2018.

\bibitem{hui2019lightweight}
Zheng Hui, Xinbo Gao, Yunchu Yang, and Xiumei Wang.
\newblock Lightweight image super-resolution with information
  multi-distillation network.
\newblock In {\em ACMMM}, 2019.

\bibitem{hwang2021layer}
Jiwon Hwang, A.~F. M.~Shahab Uddin, and Sung-Ho Bae.
\newblock A layer-wise extreme network compression for super resolution.
\newblock {\em IEEE Access}, 2021.

\bibitem{jiang2021training}
Xinrui Jiang, Nannan Wang, Jingwei Xin, Keyu Li, Xi Yang, and Xinbo Gao.
\newblock Training binary neural network without batch normalization for image
  super-resolution.
\newblock In {\em AAAI}, 2021.

\bibitem{jung2019learning}
Sangil Jung, Changyong Son, Seohyung Lee, Jinwoo Son, Jae-Joon Han, Youngjun
  Kwak, Sung~Ju Hwang, and Changkyu Choi.
\newblock Learning to quantize deep networks by optimizing quantization
  intervals with task loss.
\newblock In {\em CVPR}, 2019.

\bibitem{kim2019fine}
Heewon Kim, Seokil Hong, Bohyung Han, Heesoo Myeong, and Kyoung~Mu Lee.
\newblock Fine-grained neural architecture search.
\newblock {\em arXiv preprint arXiv:1911.07478}, 2019.

\bibitem{kim2016accurate}
Jiwon Kim, Jungkwon Lee, and Kyoung~Mu Lee.
\newblock Accurate image super-resolution using very deep convolutional
  networks.
\newblock In {\em CVPR}, 2016.

\bibitem{krishnamoorthi2018quantizing}
Raghuraman Krishnamoorthi.
\newblock Quantizing deep convolutional networks for efficient inference: A
  whitepaper.
\newblock {\em arXiv preprint arXiv:1806.08342}, 2018.

\bibitem{ledig2017photo}
Christian Ledig, Lucas Theis, Ferenc Husz{\'a}r, Jose Caballero, Andrew
  Cunningham, Alejandro Acosta, Andrew Aitken, Alykhan Tejani, Johannes Totz,
  Zehan Wang, and Wenzhe Shi.
\newblock Photo-realistic single image super-resolution using a generative
  adversarial network.
\newblock In {\em CVPR}, 2017.

\bibitem{Li2020PAMSQS}
Huixia Li, Chenqian Yan, Shaohui Lin, Xiawu Zheng, B. Zhang, F. Yang, and
  Rongrong Ji.
\newblock Pams: Quantized super-resolution via parameterized max scale.
\newblock In {\em ECCV}, 2020.

\bibitem{bee2017enhanced}
Bee Lim, Sanghyun Son, Heewon Kim, Seungjun Nah, and Kyoung~Mu Lee.
\newblock Enhanced deep residual networks for single image super-resolution.
\newblock In {\em CVPRW}, 2017.

\bibitem{lin2016fixed}
Darryl Lin, Sachin Talathi, and Sreekanth Annapureddy.
\newblock Fixed point quantization of deep convolutional networks.
\newblock In {\em ICML}, 2016.

\bibitem{lou2019autoq}
Qian Lou, Feng Guo, Lantao Liu, Minje Kim, and Lei Jiang.
\newblock Autoq: Automated kernel-wise neural network quantization.
\newblock {\em arXiv preprint arXiv:1902.05690}, 2019.

\bibitem{ma2019efficient}
Yinglan Ma, Hongyu Xiong, Zhe Hu, and Lizhuang Ma.
\newblock Efficient super resolution using binarized neural network.
\newblock In {\em CVPRW}, 2019.

\bibitem{miyashita2016convolutional}
Daisuke Miyashita, Edward~H Lee, and Boris Murmann.
\newblock Convolutional neural networks using logarithmic data representation.
\newblock {\em arXiv preprint arXiv:1603.01025}, 2016.

\bibitem{nagel2019data}
Markus Nagel, Mart~van Baalen, Tijmen Blankevoort, and Max Welling.
\newblock Data-free quantization through weight equalization and bias
  correction.
\newblock In {\em ICCV}, 2019.

\bibitem{nagel2021white}
Markus Nagel, Marios Fournarakis, Rana~Ali Amjad, Yelysei Bondarenko, Mart van
  Baalen, and Tijmen Blankevoort.
\newblock A white paper on neural network quantization, 2021.

\bibitem{medical2}
M.~Dirk Robinson, Stephanie~J. Chiu, Cynthia~A. Toth, Joseph~A. Izatt,
  Joseph~Y. Lo, and Sina Farsiu.
\newblock New applications of super-resolution in medical imaging.
\newblock {\em Digital Imaging and Computer Vision, CRC Press.}, 2010.

\bibitem{shi2016realtime}
Wenzhe Shi, Jose Caballero, Ferenc Huszar, Johannes Totz, Andrew~P. Aitken, Rob
  Bishop, Daniel Rueckert, and Zehan Wang.
\newblock Real-time single image and video super-resolution using an efficient
  sub-pixel convolutional neural network.
\newblock In {\em CVPR}, 2016.

\bibitem{Wang2021fully}
Hu Wang, Peng Chen, Bohan Zhuang, and Chunhua Shen.
\newblock Fully quantized image super-resolution networks.
\newblock In {\em ACMMM}, 2021.

\bibitem{wang2019learning}
Ziwei Wang, Jiwen Lu, Chenxin Tao, Jie Zhou, and Qi Tian.
\newblock Learning channel-wise interactions for binary convolutional neural
  networks.
\newblock In {\em CVPR}, 2019.

\bibitem{xin2020binarized}
Jingwei Xin, Nannan Wang, Xinrui Jiang, Jie Li, Heng Huang, and Xinbo Gao.
\newblock Binarized neural network for single image super resolution.
\newblock In {\em ECCV}, 2020.

\bibitem{zhang2018lq}
Dongqing Zhang, Jiaolong Yang, Dongqiangzi Ye, and Gang Hua.
\newblock Lq-nets: Learned quantization for highly accurate and compact deep
  neural networks.
\newblock In {\em ECCV}, 2018.

\bibitem{satellite2}
H. Zhang, Z. Yang, L. Zhang, and H. Shen.
\newblock Super resolution reconstruction for multi-angle remote sensing images
  considering resolution differences.
\newblock {\em Remote Sensing}, 2014.

\bibitem{military}
L. Zhang, H. Zhang, H. Shen, and P. Li.
\newblock A super resolution reconstruction algorithm for surveillance images.
\newblock {\em Signal Processing}, 2010.

\bibitem{zhang2018image}
Yulun Zhang, Kunpeng Li, Kai Li, Lichen Wang, Bineng Zhong, and Yun Fu.
\newblock Image super-resolution using very deep residual channel attention
  networks.
\newblock In {\em ECCV}, 2018.

\bibitem{zhang2018residual}
Yulun Zhang, Yapeng Tian, Yu Kong, Bineng Zhong, and Yun Fu.
\newblock Residual dense network for image super-resolution.
\newblock In {\em CVPR}, 2018.

\bibitem{zhao2019improving}
Ritchie Zhao, Yuwei Hu, Jordan Dotzel, Christopher De~Sa, and Zhiru Zhang.
\newblock Improving neural network quantization without retraining using
  outlier channel splitting.
\newblock {\em arXiv preprint arXiv:1901.09504}, 2019.

\bibitem{zhou2016dorefa}
Shuchang Zhou, Yuxin Wu, Zekun Ni, Xinyu Zhou, He Wen, and Yuheng Zou.
\newblock Dorefa-net: Training low bitwidth convolutional neural networks with
  low bitwidth gradients.
\newblock {\em arXiv preprint arXiv:1606.06160}, 2016.

\end{thebibliography}
}

\end{document}


\onecolumn
\title{
DAQ: Channel-Wise Distribution-Aware Quantization for Deep Image Super-Resolution Networks \\
-- \textit{Supplementary Document} -- }

\makeatletter
\g@addto@macro\@maketitle{
    
}
\makeatother

\maketitle
\thispagestyle{empty}

\section{Implementation details on computation cost}
\label{sup:overflow}

In the main paper, computational resources (BOPs and estimated energy consumption) are measured with respect to quantization bit-width, considering the overflow of low-bit arithmetic operations.
Overflow occurs when an arithmetic operation attempts to create a numeric value that is outside the range that can be possibly represented.
Taking integer overflow into account is especially essential in low-bit networks,
since the output ranges of low-bit multiplication and addition are strictly limited.
Various techniques are used to avoid the integer overflow, such as using the overflow checker or value sanity testing.
For ultra-low precision operations where the integer overflow is highly likely to occur, we design an appropriate large bit-width for each operation, under the assumption of an integer overflow.
For instance, when the sum is accumulated over vector of size $C$ with each $n$-bit element, the output buffer is $n\Plus log_2{(C\Minus 1)}$-bit.
Also, the output buffer for multiplication of two $n$-bit elements is $(2n\Minus1)$-bit.

\section{Derivation of convolution operations for DAQ}
\label{sup:derivation}

Section~{\textcolor{red}{3.3}} in the main paper claims that quantization can step forward to hardware efficiency simply by postponing the de-transformation process as late as possible (See Equation~\eqref{eqn:sup-convb},~\eqref{eqn:sup-convd}, and ~\eqref{eqn:sup-qq-g}).
As more operations are done before de-transformation, in other words, in the state of real integer, the more efficient the quantization becomes.

Section~{\textcolor{red}{3.2}} of the main paper presents a convolution operation with a $n$-bit \textit{channel-wise} quantized feature map and a $n$-bit layer-wise quantized weight tensor.
Given a feature map $\boldsymbol{x} \in \mathbb{R}^{C\times H\times W}$,
$c$-th channel, $j, k$-th element of the feature map is denoted as ${x}_c[j, k]$
with the indexing operator $[\cdot,\cdot]$.
Given a weight $\boldsymbol{w} \in \mathbb{R}^{C\times C_{out}\times K\times K}$,
c-th input channel and $i, u, v$-th element of a part of weight tensor $\boldsymbol{w} \in \mathbb{R}^{C\times C_{out}\times K\times K}$ is denoted as ${w_c}[i,u,v]$ with indexing operator $[\cdot,\cdot,\cdot]$.
Then, the output response $\boldsymbol{y} \in \mathbb{R}^{C_{out}\times H\times W}$ is the output of convolution between the given feature map and the weight in a sliding window manner. The $i, j, k$-th element of the output response is formulated as follows:
\begin{equation}
    y[i,j,k]=\sum_{c=1}^{C}\sum_{u=1}^{K}\sum_{v=1}^{K}{  x_c[u\Plus j,v\Plus k] \cdot {w}_c[i,u,v] }.
\label{eqn:sup-conva}
\end{equation}
For simplicity, we drop the index subscript in the following equations to denote 
$x_c[u\Plus j,v\Plus k]$ as $x_c$ and ${w}_c[i,u,v]$ as $w_c$.
From our proposed quantization method, convolution with floating-point values can be approximated with low-precision values, as follows:
\begin{equation}
    y[i,j,k]\approx
    \sum_{c=1}^{C}\sum_{u=1}^{K}\sum_{v=1}^{K}{ x^q_c \cdot {w}^q_c }
\label{eqn:sup-convb}
\end{equation}

\begin{equation}
    =\sum_{c=1}^{C}\sum_{u=1}^{K}\sum_{v=1}^{K}{ (\s_{c}s(n)\cdot \hat{x}^q_c + \mu_{c})\cdot(\s_w{s(n)}\cdot \hat{w}^q_c) }
\label{eqn:sup-convc}
\end{equation}

\begin{equation}
    ={\s_w}s(n)^2 \sum_{c=1}^{C}\s_{c}\cdot \sum_{u=1}^{K}\sum_{v=1}^{K}{ \hat{x}^q_c\cdot\hat{w}^q_c }
    +{\s_w}s(n) \sum_{c=1}^{C} {\mu_c} \sum_{u=1}^{K}\sum_{v=1}^{K}{\hat{w}^q_c }.
\label{eqn:sup-convd}
\end{equation}
Likewise overview Figure~{\textcolor{red}{3}} in the main paper, the channel-wise de-transformation (see Equation~\eqref{eqn:sup-convd}) derived from the procedure with element-wise de-transformation (see Equation~\eqref{eqn:sup-convb}) can reduce costly operations with floating-point values.
Although the computation costly operation of element-wise de-transformation is largely reduced in Equation~\eqref{eqn:sup-convd},
it still bears computational overhead, by operating the channel-wise summation in floating-point values (due to floating-point de-transformation parameters $\mu_c$ and $\s_c$).
To alleviate this issue, main paper presents a scheme of quantizing quantization transformation parameters of $\boldsymbol\mu\in \mathbb{R}^{C}$ and $\boldsymbol\s\in \mathbb{R}^{C}$, to approximate $\mu_c$ and $\s_c$ by using the distribution statistics of $\boldsymbol\mu$ and $\boldsymbol\s$.
From Equation~{\textcolor{red}{4}, \textcolor{red}{5}, \textcolor{red}{6}} of the main paper,
Equation~\eqref{eqn:sup-convd} can be approximated as follows:
\begin{equation}
    \approx{\s_w}s(n)^2 \sum_{c=1}^{C}\textcolor{blue}{\s_{c}^q}\cdot \sum_{u=1}^{K}\sum_{v=1}^{K}{ \hat{x}^q_c\cdot\hat{w}^q_c}
    +{\s_w}s(n) \sum_{c=1}^{C} \textcolor{blue}{\mu_c^q} \sum_{u=1}^{K}\sum_{v=1}^{K}{\hat{w}^q_c}
\label{eqn:sup-qq-e}
\end{equation}

\begin{equation}
    ={\s_w}s(n)^2 \sum_{c=1}^{C}\textcolor{blue}{(\s_{\s}s(m)\cdot\hat{\s}^q_c+\mu_\s)}\cdot \sum_{u=1}^{K}\sum_{v=1}^{K}{ \hat{x}^q_c\cdot\hat{w}^q_c }
    +{\s_w}s(n) \sum_{c=1}^{C} \textcolor{blue}{(\s_{\mu}s(m)\cdot\hat{\mu}^q_c+\mu_\mu)} \sum_{u=1}^{K}\sum_{v=1}^{K}{\hat{w}^q_c }
\label{eqn:sup-qq-f}
\end{equation}

\begin{equation}
\begin{multlined}
    ={\s_w}s(n)^2 \s_{\s}s(m)\sum_{c=1}^{C}\hat{\s}^q_c\sum_{u=1}^{K}\sum_{v=1}^{K}{ \hat{x}^q_c\cdot\hat{w}^q_c }
    +{\s_w}s(n)^2 \mu_\s \sum_{c=1}^{C}\sum_{u=1}^{K}\sum_{v=1}^{K}{ \hat{x}^q_c\cdot\hat{w}^q_c } \\
    +{\s_w}s(n)\s_{\mu}s(m)\sum_{c=1}^{C} {\hat{\mu}^q_c} \sum_{u=1}^{K}\sum_{v=1}^{K}{\hat{w}^q_c }
    +{\s_w}s(n)\mu_\mu \sum_{c=1}^{C} \sum_{u=1}^{K}\sum_{v=1}^{K}{\hat{w}^q_c }.
\end{multlined}
\label{eqn:sup-qq-g}
\end{equation}
The floating-point channel-wise summation (see Equation~\eqref{eqn:sup-convd}) is replaced with lower-precision channel-wise summation (see Equation~\eqref{eqn:sup-qq-g}).
As shown in Table~\ref{tab:sup-bops}, simply changing the operation order from Equation~\eqref{eqn:sup-conva} to Equation~\eqref{eqn:sup-convd} reduces the BOPs largely from 174T to 10T. 
Furthermore, \textbf{q}uantizing \textbf{q}uantization transformation parameters (QQ) in Equation~\eqref{eqn:sup-convd} results in Equation~\eqref{eqn:sup-qq-g}, which further reduces the BOPs to 3T.
\begin{table*}[!htbp]
\caption{Computational cost comparison of a 2-bit (w2a2) channel-wise quantized convolution. $C\Equal C_{out}\Equal 256$, $K\Equal3$, $(H,W)\Equal(480,270)$}
\centering
\scalebox{0.9}{
\begin{tabular}{cc c cccc cccr r}
\toprule
\multicolumn{2}{c}{De-transformation} & \multirow{3}{*}{Eqn.} & \multicolumn{8}{c}{Number of ($n$-bit, $m$-bit) operations} &  \multirow{3}{*}{BOPs} \\
\cmidrule(lr){1-2} \cmidrule(lr){4-11} 
Type & QQ & & (2, 2) & (3, 3) & (6, 4) & (6, 6) & (9, 9) & (14, 32) & (17, 32) & (32, 32) &  \\ 
\midrule 
Element-wise & \xm & Eqn.~\eqref{eqn:sup-convb} & - & - & - & - & - & - & - & 169937M & 174015G\\
\cmidrule(lr){1-12} 
Channel-wise & \xm & Eqn.~\eqref{eqn:sup-convd}& 76441M & 67948M & - & 8493M & - & 8493M & - & 8493M & 10108G\\
\cmidrule(lr){1-12} 
Channel-wise & \cm &Eqn.~\eqref{eqn:sup-qq-g}& 152882M & 135895M & 8493M & 8493M & 8460M & 33M & 33M & 66M & 3046G\\
\bottomrule
\end{tabular}
}
\label{tab:sup-bops}
\end{table*}

\clearpage
\section{Additional experiments} 
\subsection{Comparison with SotA methods}

Existing state-of-the-art quantized SR networks~[\textcolor{green}{23}, \textcolor{green}{41}] involve a specialized architecture or an ad-hoc training scheme for low precision SR networks, mostly concentrated on binary precision. 
BTM~[\textcolor{green}{23}] exploits a new training scheme like knowledge distillation and specialized gradient update rule instead of the traditional straight-through estimator~[\textcolor{green}{46}].
BAM~[\textcolor{green}{41}] designs a new binarized SR network, namely BSRN, utilizing a bit accumulation module.

Our proposed quantization method is orthogonal to these techniques.
EDSR-BTM and BSRN-BAM are re-implemented according to the respective paper, and we replace the typical quantization (binarization) function with our distribution-aware channel-wise quantization (DAQ) function.
The re-implemented architectures and the DAQ-applied architectures are respectively trained with batch size 4 and other settings same as the baseline in each paper.
Despite the channel-wise overhead, our proposed method DAQ gives clear auxiliary gain in performance, for about 0.3 dB in Set5, as shown in Table~\ref{tab:sup-exp}.

\begin{table*}[h!]
\caption{ Comparisons on existing low-precision SR networks, EDSR-BTM and BSRN-BAM of scale 4.
}
\vspace{-0.2cm}
\centering
\scalebox{0.95}{
\begin{tabular}{l cc rrr cccc}
\toprule
\multirow{2}{*}{Method} &\multicolumn{2}{c}{Precision}& \multirow{2}{*}{BOPs}& \multirow{2}{*}{Energy}& \multirow{2}{*}{Parameters}&\multicolumn{4}{c}{PSNR (dB)} \vspace{-0.1cm}\\
\cmidrule(lr){2-3}  \cmidrule(lr){7-10}  
\multicolumn{1}{c}{}& w& a&&&&Set5 & Set14 & B100 & Urban100 \\
\midrule
EDSR-BTM~[\textcolor{green}{23}] &1&1&23.1 T&52.8 mJ&43.1M& 31.30&28.05&27.22&25.08\\
EDSR-BTM~[\textcolor{green}{23}] - DAQ &1&1& 75.1 T	&138.9 mJ &43.1M &31.60&28.19&27.34&25.30\\
\midrule
BSRN-BAM~[\textcolor{green}{41}] &1&1& 2.9 T&8.5 mJ&1.2M &31.17&27.94&27.15&25.01\\
BSRN-BAM~[\textcolor{green}{41}] - DAQ &1&1& 7.2 T & 23.7 mJ &1.2M
&31.44&28.03&27.21&25.05\\
\bottomrule
\end{tabular}}
\label{tab:sup-exp}
\end{table*}

\subsection{SR Networks with batch normalization layers}
In the main paper, we made a comparison with quantization methods without retraining.
Among the compared methods, DFQ~{[\textcolor{green}{37}]} utilizes batch normalization (BN) parameters to further improve the quantization accuracy.
However, the comparison backbone of Table~{\textcolor{red}{3}},
EDSR~{[\textcolor{green}{31}]} removed the BN layers for improved performance, followed by several other SR networks.
For further fair comparison with DFQ, we compare the quantization methods without retraining, on EDSR \textit{with BN}.
Table~\ref{tab:sup-bn} shows that our method outperforms DFQ regardless of BN layers.

\begin{table}[!htbp]
\caption{Comparison of quantization methods \textit{without retraining} on pre-trained EDSR \textit{with BN} of scale 4. }
\vspace{-0.2cm}
\centering
\scalebox{0.9}{
\begin{tabular}{l cc rrr }
    \toprule
    \multirow{2}{*}{Method} & \multicolumn{2}{c}{Precision} & \multirow{2}{*}{BOPs} & \multirow{2}{*}{Energy} & \multirow{2}{*}{PSNR} \\
    \cmidrule(lr){2-3} 
    & w& a & (HD image) & (HD image) & (Urban100) \\ 
    \midrule
    EDSR w/ BN & 32 & 32 & 10025.8 T& 22516.0 mJ& 26.04 dB\\
    \midrule
    EDSR w/ BN - LinQ & 4 & 4 & 357.8 T& 378.4 mJ& 22.79 dB\\
    EDSR w/ BN - DFQ  & 4 & 4 & 364.3 T& 390.1 mJ& 23.07 dB\\ \rowcolor{orange!25}
    EDSR w/ BN - DAQ  & 2 & 2 & 213.4 T& 333.5 mJ& \textbf{24.53 dB} \\
    \bottomrule
\end{tabular}
}
\label{tab:sup-bn}
\end{table}